\documentclass[letterpaper, 10 pt, conference]{ieeeconf}  

\IEEEoverridecommandlockouts                              

\usepackage[utf8]{inputenc}
\usepackage[T1]{fontenc}
\usepackage{lmodern}
\usepackage{anyfontsize}
\usepackage{microtype}
\usepackage{subcaption}
\usepackage{algorithm}
\usepackage{algorithmic}
\usepackage{indentfirst}
\usepackage{multicol}
\usepackage{url}

\usepackage[font=small,labelfont=bf,skip=2pt]{caption}
\usepackage{dutchcal}

\usepackage{conventions}

\usepackage{pgfplots}
\DeclareUnicodeCharacter{2212}{−}
\usepgfplotslibrary{groupplots,dateplot}
\usetikzlibrary{patterns,shapes.arrows}
\pgfplotsset{compat=newest}

\tikzset{every mark/.append style={mark size=1pt}}

\usepackage[capitalise, noabbrev]{cleveref}

\usepackage{todonotes}
\usepackage{xcolor}

\makeatletter
\newcommand{\printfnsymbol}[1]{%
  \textsuperscript{\@fnsymbol{#1}}%
}
\makeatother

\title{\LARGE \bf Fusing RGBD Tracking and Segmentation Tree Sampling for Multi-Hypothesis Volumetric Segmentation}
\author{Andrew Price\textsuperscript{*} \and Kun Huang\textsuperscript{*} \and Dmitry Berenson

\thanks{*Andrew Price and Kun Huang contributed equally to this work. This work was supported in part by NSF grant IIS-1750489 and by Toyota Research Institute (TRI). This article solely reflects the opinions of its authors and not TRI or any other Toyota entity. The authors are with the University of Michigan, Ann Arbor, MI, USA.{\tt\small \{huangkun, pricear, dmitryb\}@umich.edu}}%
}
\begin{document}

\clearpage \pagebreak

\maketitle
\global\csname @topnum\endcsname 0
\global\csname @botnum\endcsname 0

\begin{abstract}
Despite rapid progress in scene segmentation in recent years, 3D segmentation methods are still limited when there is severe occlusion. The key challenge is estimating the segment boundaries of (partially) occluded objects, which are inherently ambiguous when considering only a single frame. In this work, we propose Multihypothesis Segmentation Tracking (MST), a novel method for volumetric segmentation in changing scenes, which allows scene ambiguity to be tracked and our estimates to be adjusted over time as we interact with the scene. Two main innovations allow us to tackle this difficult problem: 1) A novel way to sample possible segmentations from a segmentation tree; and 2) A novel approach to fusing tracking results with multiple segmentation estimates. These methods allow MST to track the segmentation state over time and incorporate new information, such as new objects being revealed.
We evaluate our method on several cluttered tabletop environments in simulation and reality. Our results show that MST outperforms baselines in all tested scenes.
\end{abstract}

\section{Introduction}

Instance segmentation of 3D environments is a crucial problem for robotic manipulation, particularly in tabletop and household scenarios.
Fortunately, instance segmentation of images and videos has made rapid progress in recent years, driven by advances in computational capacity, dataset size, and learning algorithm innovations.
Even so, current state-of-the-art algorithms \cite{pham2018scenecut,xie2020best} will struggle with moderately-complex scenes that are common in manipulation-in-clutter tasks.
Similarly, modern semi-supervised video segmentation \cite{caelles2017osvos,wang2019fast} will be limited by the quality of the initial masks, placing increasing importance on the quality of a single-frame segmentation.
Instance segmentation in 3D is even more challenging, requiring either multiview mapping or volumetric shape completion, which are still active areas of research \cite{xu2019midfusion,venkataraman2019kinematically,schiebener2013integrating,tulsiani2018factoring,lundell2019robust}.

3D segmentation in cluttered scenes is challenging for several reasons. First, occlusions from objects in the environment and the robot itself can be severe, even entirely occluding some objects. Second, human environments can be highly dynamic and greatly varied, meaning that most scenes will be novel to some degree, precluding the use of model-registration techniques \cite{icp}. While using appropriate priors, high-fidelity sensing and memory, and a physics-based constraints can be helpful, an effective approach also requires a way to explicitly reason about uncertainty in segmentation estimates.

\begin{figure}
\setlength{\belowcaptionskip}{0pt}
\centering
    \begin{subfigure}[b]{.3\linewidth}
        \includegraphics[height=2cm]{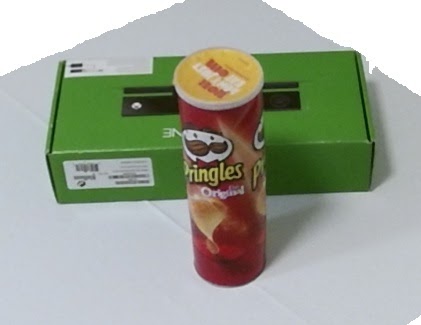}
        \caption{RGB at $t=2$}
    \end{subfigure}
    \begin{subfigure}[b]{.33\linewidth}
        \includegraphics[height=2cm]{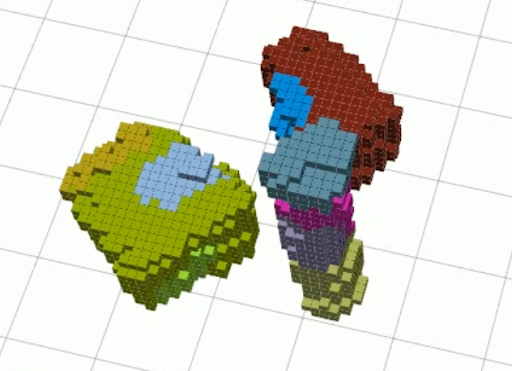}
        \caption{\cite{price2019inferring}}
    \end{subfigure}
    \begin{subfigure}[b]{.33\linewidth}
        \includegraphics[height=2cm]{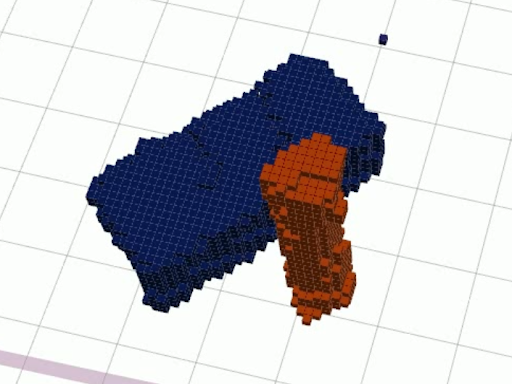}
        \caption{MST}
    \end{subfigure}

    \begin{subfigure}[b]{.3\linewidth}
        \includegraphics[height=1.8cm]{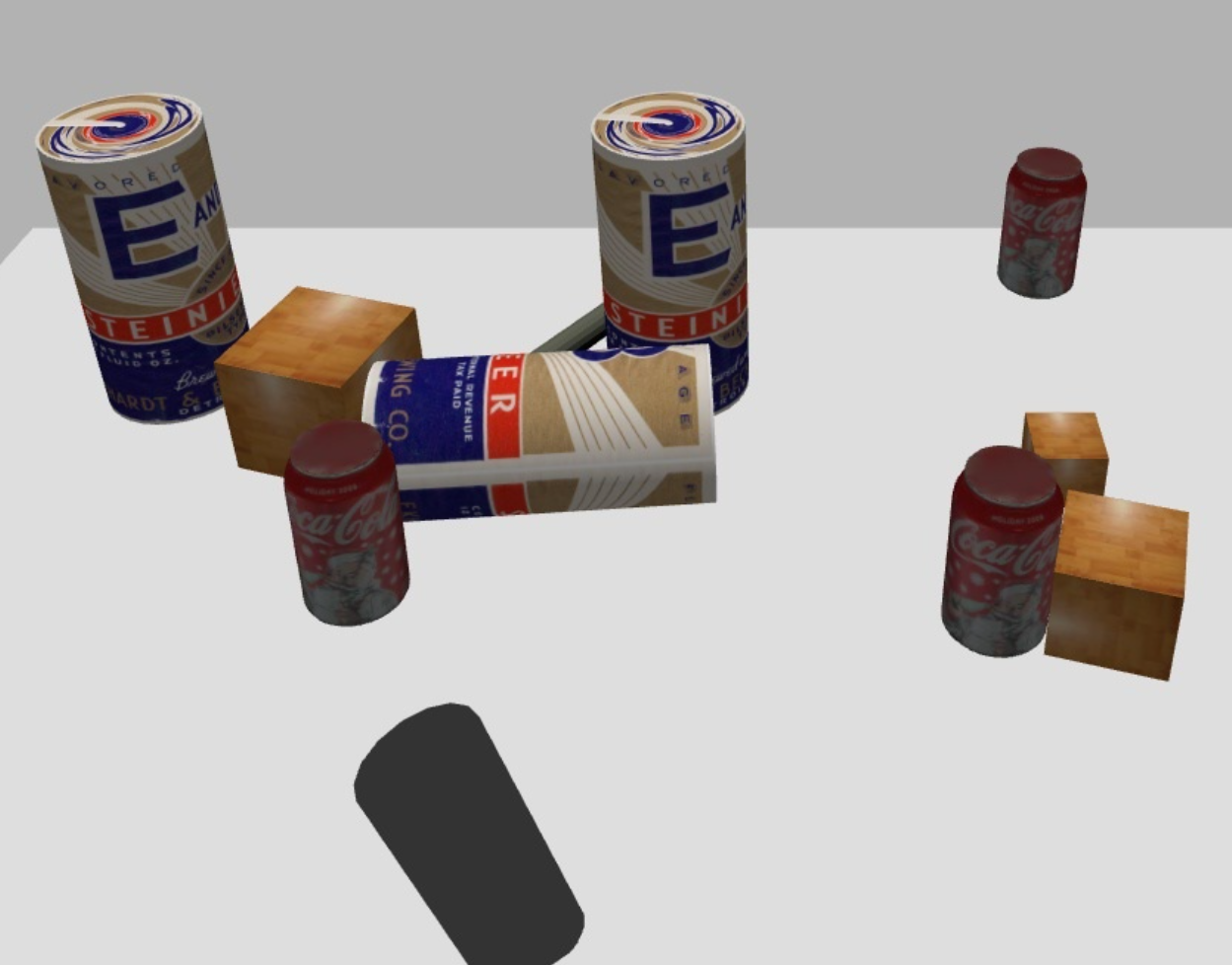}
        \caption{RGB at $t=2$}
    \end{subfigure}
    \begin{subfigure}[b]{.33\linewidth}
        \includegraphics[height=1.8cm]{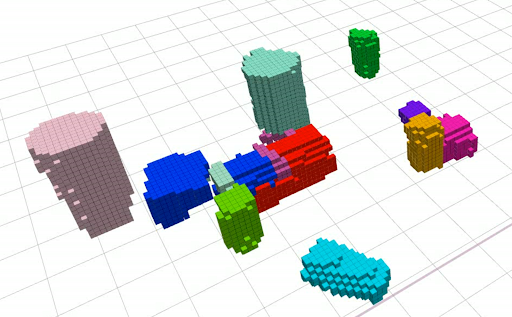}
        \caption{\cite{price2019inferring}}
    \end{subfigure}
    \begin{subfigure}[b]{.33\linewidth}
        \includegraphics[height=1.8cm]{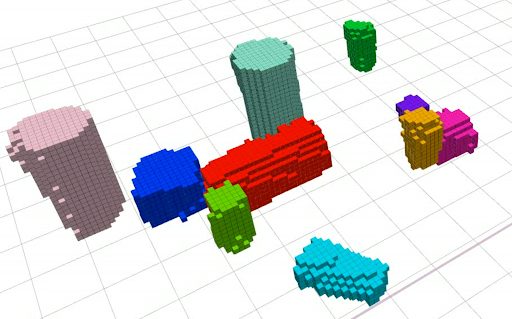}
        \caption{MST}
    \end{subfigure}
    \setlength{\belowcaptionskip}{-15pt}
\caption{Qualitative comparison of the volumetric segmentation results obtained directly using SceneCut \cite{pham2018scenecut} + \cite{price2019inferring} and our method (MST) on both real-world and simulation experiments.}
\label{fig:teaser}
\end{figure}

Multi-hypothesis state representations are commonly used in robot navigation to deal with uncertain states and measurements, but they are seldom employed in segmentation. Maintaining a set of hypotheses about a scene can be useful for many manipulation applications, e.g. planning actions which are robust to uncertainty or for active perception. Most importantly, maintaining a set of hypotheses enables an algorithm to consider possibilities which are not currently the most probable, but may be a better basis for estimates when future data is received.

In this work, we present, Multihypothesis Segmentation Tracking (MST), a novel fusion of sampling and tracking methods to perform multi-hypothesis volumetric instance segmentation of cluttered scenes.
Specifically, our contributions are 1) A novel Markov Chain Monte Carlo (MCMC) technique for sampling plausible segmentations from a segmentation tree; and 2) A novel approach to fusing tracking and segmentation measurements for multiple segmentation hypotheses.
These methods allow us to maintain multiple uncertain but plausible segmentations across time and to incorporate new information, such as a new object being revealed.

Our experiments\footnote{\url{https://youtu.be/kottSLebgBA}} show that MST outperforms previous work employing single-instant scene segmentation \cite{price2019inferring} and using video object tracking \cite{wang2019fast} in generating 3D segmentations. Our code is available open-source\footnote{\url{https://github.com/UM-ARM-Lab/multihypothesis_segmentation_tracking}}.

\begin{figure*}
\includegraphics[width=\linewidth]{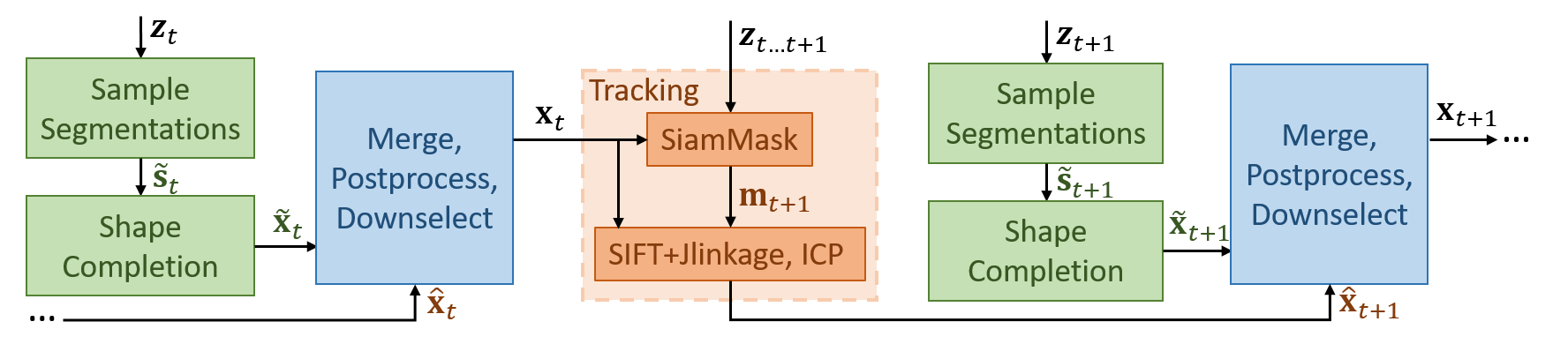}
\caption{An overview of the main components of our method. Green: segmentation sampling; Blue: Merging and processing estimates; Orange: Tracking. $\indexed{\predicted{\particle}}{}{}{0} = \emptyset$.}
\label{fig:blockdiagram}
\end{figure*}

\section{Related Work} \label{sec:related}
Instance segmentation of manipulation scenes is a well-studied and rich field \cite{bohg2017interactive}.
However, some limitations occur regularly, including framing the segmentation problem in 2D or 2.5D space \cite{van2014probabilistic} (i.e. not estimating full volumetric occupancy), relying on pre-specified \cite{sui2017sum} or simple geometric models \cite{novkovic2019object} in the scene, or restricting the belief about the scene to a unimodal representation, even if tracking is performed in a multimodal fashion \cite{hausman2013tracking}.
Working in 3D, as opposed to 2D or 2.5D, is particularly important, as it allows us to construct and retain object geometry estimates in the presence of occlusion, which is frequent in cluttered scenes.
In this work, we aim to embrace the challenges of manipulating unknown objects in real environments by addressing the problem in its native 3D occupancy space, acknowledging that our segmentation and tracking tools are imperfect by considering multiple hypotheses of the segmentation state.

Scene segmentation has been addressed in a variety of ways, and though we do not attempt a complete taxonomy here, several broad classes of techniques can be found in the literature. 
With the development of deep learning object detection, some approach the instance segmentation problem by treating it as object detection \cite{he2017mask, wang2019solo}.
Another approach is to localize known objects within the scene, providing 3D knowledge from prior geometric information \cite{xiang2017posecnn,zeng2017multi,schwarz2018rgb}.
While very powerful in certain contexts, these two approaches scale poorly to handling general household manipulation scenes with novel object categories.

With the growth in deep networks over the past decade, there has been significant work in producing an occupancy grid from a single depth image or a stitched 3D scene
\cite{Hou2019CVPR, Wu_2015_CVPR,Maturana2015VoxNet,Qi_2016_CVPR,Riegler_2017_CVPR,Wang_2019_ICCV,Meng2019ICCV}.
While there are some similarities between this problem and ours, there are also significant differences: these methods operate on sequences of images of static scenes (as opposed to the dynamic ones we consider), and while the outputs may have an associated probability, they do not provide multiple segmentation hypotheses, as we aim to do here.

One application domain that has seen significant work on voxel-based segmentation and classification is medical imaging, since techniques like MRI or CT offer true 3D imaging instead of the 2.5D of RGBD sensors. Uses include mapping airways \cite{Lo2010Vessel} and brain tissue \cite{Vanderlijn2008Hippocampus, Quadrelli2016Hitchhikers,Nair2020Uncertainty}, with techniques like graph cuts, KNN, and deep networks drawn from the broader perception community, plus hand-tuned heuristic ones drawn from biology domain knowledge. In our work, we do not assume true 3D information is available and must rely on RGBD images.

Related work in sensor fusion via particle filter, which has been commonly used for non-linear state estimation, has seen various improvements on sampling sufficient valid states and avoiding degeneracy of the proposal distribution \cite{koval2017manifold, kwok2005evolutionary}. Although these probabilistic methods have been used in manipulation \cite{van2014probabilistic, sui2017sum}, they either only produce 2D estimates, or require prior knowledge of object models.

\begin{figure*}[t]
\centering
\includegraphics[width=\linewidth]{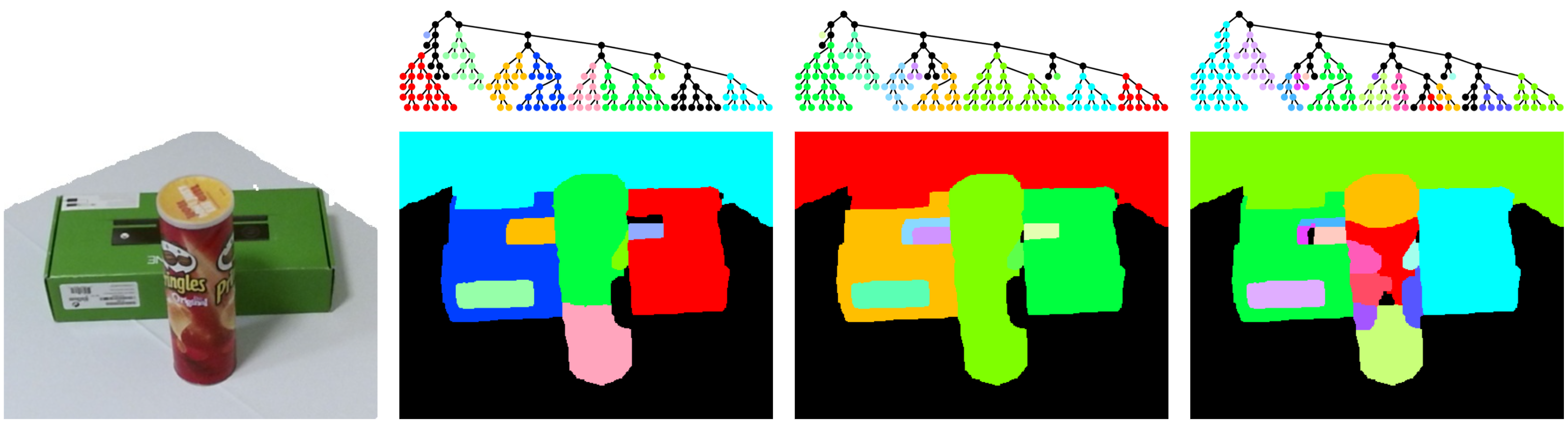}
\caption{Sampled tree cuts and their resulting image segmentations. The segmentation trees have been truncated for clarity; the full size is approximately 1000 nodes.}
\label{fig:seg_samples}
\end{figure*}

\section{Problem Statement}

Let $\voxelIndex \in \N^3$ denote the coordinate of one voxel. $\objectCount \in \N$ is the estimated number of objects in the region of interest.
Let  $\objectIndex \in \objectIndexSpace \defined \setof{0,1,2,...,\objectCount}$ denote the object label of $v$ where $k=0$ means the voxel is in free space.
Our segmentation state vector $\particle$ is an assignment from a voxel coordinate to an object label.
Similarly, an image segmentation $\segmentation$ assigns a label to each pixel.
\begin{equation}
    \particle \from \voxelIndexSpace \to \objectIndexSpace
    , \hspace{5em}
    \segmentation \from \pixelIndexSpace \to \objectIndexSpace
\end{equation}

For a multi-hypothesis system evolving over discretized time, we will use $\timeIndex \in \setof{1,...,\timeCount}$ to indicate the time index, $\hypothesisIndex \in \setof{1,...,\hypothesisCount}$ to indicate the hypothesis index, and $k$ to indicate an object index.
So, a single object hypothesis at a point in time can be written as $\objectFull \defined \setof{\voxelIndex \in \voxelIndexSpace \st \particleFull[v]{i}{t} = k}$.

To compute the similarity of two segmentations, we can define a match quality $\quality \from \stateSpace \times \stateSpace \to \rangeof{0, 1}$ where $q=1$ represents a perfect match between two segmentation states.
For this work, use a symmetric version of the weighted coverage \cite{silberman2014instance}:
\begin{align}
    \quality(\indexed{\particle}{}{i}{}, \indexed{\particle}{}{j}{})
    &\defined \tfrac{1}{2}C(\indexed{\particle}{}{i}{}, \indexed{\particle}{}{j}{})+\tfrac{1}{2}C(\indexed{\particle}{}{j}{}, \indexed{\particle}{}{i}{}), \label{eq:quality_fn}
    \\
    C(\indexed{\particle}{}{i}{}, \indexed{\particle}{}{j}{})
    &\defined
    \tfrac{1}{\card{\voxelIndexSpace}}
    \sum_{m \in \indexed{\objectIndexSpace}{}{i}{}}
    \card{\indexed{\object}{m}{i}{}}
    \max_\alpha \iou(\indexed{\object}{m}{i}{}, \indexed{\object}{\alpha}{j}{})
\end{align}
where
$
\iou(\indexed{\object}{m}{}{}, \indexed{\object}{n}{}{})
\defined \frac{\card{\indexed{\object}{m}{}{} \intersect \indexed{\object}{n}{}{}}}{\card{\indexed{\object}{m}{}{} \union \indexed{\object}{n}{}{}}}
$
represents the intersection over union (IOU, or Jaccard distance). 
The symmetric weighted coverage accounts for both false positives and false negatives in the volumetric segmentation. 
We do not use the Precision-Recall curve, which is a common metric for object detection and instance segmentation, because our goal is to segment the entire scene.

Given a sequence of RGBD images $\measurement_{0\ldots\finalTime}$ representing an object manipulation sequence, we wish to produce a set of $\hypothesisCount$ diverse 3D segmentations, which are consistent with $\measurement_{0\ldots\finalTime}$.

\section{Approach}

\subsection{Overview}

MST follows this basic outline: observe the static scene and sample possible 3D segmentations, observe an interaction with the scene and estimate the rigid motions via tracking, then combine the tracked prior segmentations with new samples directly from the subsequent static scene.
The combination of segmentation hypotheses is performed by computing which object hypotheses conflict with one another, sampling a set of merges and splits between these objects, and resampling this new population according to their current and historical fitness.
The overall process is illustrated in \cref{fig:blockdiagram} and described in detail below.

\begin{figure*}[t]
\centering
\includegraphics[width=\linewidth]{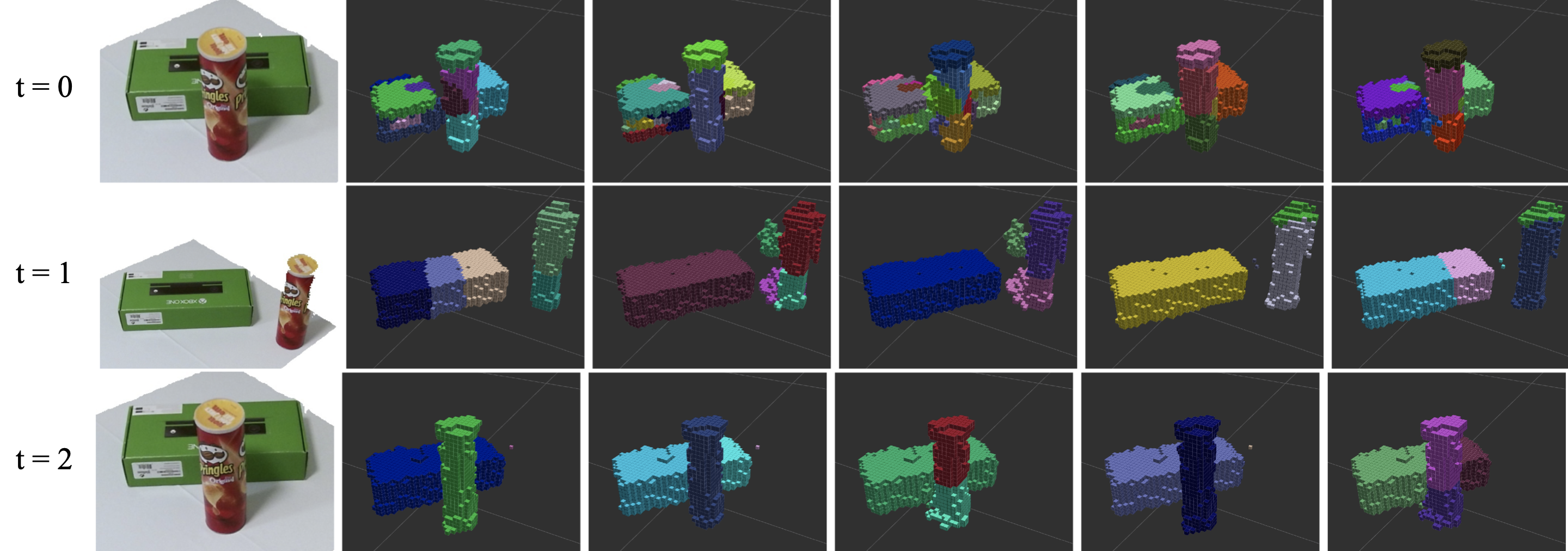}
\caption{Real-world experiment R1 with 5 hypotheses shows the convergence of hypotheses toward ground truth. Hypotheses at $t=0$ are initialized with segmentation sampling and shape completion. Our final hypotheses $\indexed{\particle}{}{}{t}$ are shown for $t=1$ and $t=2$ in decreasing order of weight $\indexed{w}{}{m}{t}$.}
\label{fig:realworld}
\end{figure*}

\begin{algorithm}[t]
\caption{Metropolis-Hastings Segmentation Sampler}
\label{alg:mhss}
\begin{algorithmic}[1]
\renewcommand{\algorithmicrequire}{\textbf{Input:}}
\renewcommand{\algorithmicensure}{\textbf{Output:}}
\REQUIRE RGBD image $\measurement$, number of samples $\hypothesisCount$, autocorrelation steps $a$
\ENSURE  Weighted sample set $\setof{\tuple{\indexed{\scenecutWeight}{}{i}{}, \indexed{\sampled{\segmentation}}{}{i}{}}}$
\STATE $\tree$ $\leftarrow$ COB($\measurement$)
\STATE $c_t$ $\leftarrow$ $c^*$ $\leftarrow$ \textsc{SceneCut}($\tree$, $\measurement$)
\FOR {$i = 1$ to $\hypothesisCount$}
\FOR {$j = 1$ to $a$}
    \STATE $k$ $\leftarrow$ \textsc{Rand}$\setof{1,\ldots,\card{c_t}}$
    \STATE $c'$ $\leftarrow$ \textsc{MoveNode}($c_t[k]$, \textsc{Rand}$\setof{\textsc{Up},\textsc{Down}}$)
    \STATE $\alpha$ $\leftarrow$ $\min(\tfrac{p(c')}{p(c_t)}\tfrac{g(c_t, c')}{g(c', c_t)},1)$
    \IF {\textsc{Rand}$(0,1) < \alpha$}
        \STATE $c_t \leftarrow c'$
    \ENDIF
\ENDFOR
\STATE $\tuple{\indexed{\scenecutWeight}{}{i}{}, \indexed{\sampled{\segmentation}}{}{i}{}} \leftarrow \tuple{p(c_t), \textsc{Apply}(c_t, \tree)}$
\ENDFOR
\RETURN $\setof{\tuple{\indexed{\scenecutWeight}{}{i}{}, \indexed{\sampled{\segmentation}}{}{i}{}}}$
\end{algorithmic}
\end{algorithm}

\subsection{Segmentation Sampling} \label{sec:sampling}

Because a single image segmentation is unlikely to be exactly correct, we wish to generate a weighted collection of segmentation hypotheses from a sensor measurement. Let the sampling procedure be defined as $\textsc{Sample} \from \indexed{\measurement}{}{}{t} \mapsto \setof{\tuple{\indexed{\scenecutWeight}{}{i}{t}, \indexed{\sampled{\segmentation}}{}{i}{t}}}$, with weight $\indexed{\scenecutWeight}{}{i}{t}$ representing the quality of the sample $i$ and operator $\sampled{\boldsymbol{\cdot}}$ indicating ``sampled''.
We begin with the segmentation tree $\tree$ (called an Ultrametric Contour Map [UCM] \cite{arbelaez2006boundary}) generated by the Convolutional Oriented Boundaries \cite{maninis2018convolutional} algorithm, which we then walk in a Metropolis-Hastings manner \cite{hastings1970monte}. 
The segmentation tree has nodes representing contiguous regions of the image, with the root node containing the whole image, and the children of a node representing a partition of that image region, such that the leaf nodes represent atomic ``superpixel'' regions.
A ``cut'' $c$ of this tree is a collection of nodes separating the root from the leaves, and representing a possible partition of the full image.
The SceneCut \cite{pham2018scenecut} algorithm assigns a value to a particular tree cut as $v(c)$, and computes the optimal cut $c^* = \max_c v(c)$.

A Metropolis-Hastings sampler is a stateful, random-walk approach to generating random samples from a distribution whose value can be computed for a given state $x$, but which can't be sampled from directly.
M-H sampling is powerful in part because it handles its inputs in a black box fashion: the details of $v(c)$ are irrelevant to the sampler as long as we can provide a the two inputs to the algorithm: the proposal distribution $g(x_t, x')$ and the posterior distribution $p(x)$.

Our M-H sampler is described in Alg. \ref{alg:mhss}. We define our M-H posterior as $p(c) \sim \exp(-(v(c^*)-v(c))^2/\sigma^2)$, and the M-H proposal distribution $g(c_t, c')$ by selecting a node in the cut $c_t$ uniformly at random and moving the cut at that node
up or down the tree to generate a proposed cut $c'$.
The $\sigma$ parameter controls how likely we are to consider lower-scoring segmentations.
We calculate the acceptance ratio $\alpha = \tfrac{p(c')}{p(c_t)}\tfrac{g(c_t, c')}{g(c', c_t)}$, and accept the proposal with probability $\min(1, \alpha)$, in the usual fashion.
The process repeats until $n$ samples have been generated, with options to insert steps for burn-in and to reduce autocorrelation.
Employing an MCMC approach means we do not need to estimate the probability of an individual node as in \cite{hu2015treecut,Snell2017StochasticST}, which is especially useful as the SceneCut results suggest that a Ultrametric Contour Map node's segmentation probability is better understood in the context of its neighbors than standalone.
To our knowledge, this the first MCMC approach for generating segmentations from a segmentation tree, and the first segmentation sampler using the SceneCut quality metric.
Some example segmentations sampled by our method are shown in \cref{fig:seg_samples}.

\subsubsection{Shape Completion and Imaging}
Given an image segmentation of the RGBD measurement, we can use shape completion $\textsc{Complete}$ \cite{price2019inferring} to estimate the occupancy of occluded voxels, which takes each $2.5$D segment as the input and reconstructs the $3$D shape using a deep neural network.
The inverse operation $\textsc{Project}$ projects a 3D segmentation to a 2D one using ray-casting (given camera intrinsics and extrinsics), and is used when evaluating scenes for which we have no 3D ground truth. 
\begin{equation}
    \textsc{Complete} \from \indexed{\sampled{\segmentation}}{}{i}{t} \mapsto \indexed{\sampled{\particle}}{}{i}{t}
    , \hspace{3em}
    \textsc{Project} \from \indexed{{\particle}}{}{i}{t} \mapsto \indexed{{\segmentation}}{}{i}{t}
\end{equation}
The completion function is run for every 2D segment in the sampled segmentations.

\begin{figure*}[t]
\includegraphics[width=\linewidth]{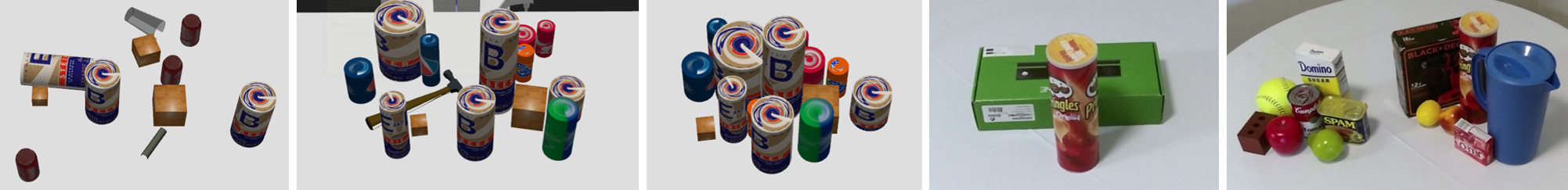}
\caption{Initial configurations of simulated experiments S1, S2, and S3, and real-world experiments R1, R2.}
\label{fig:experiments}
\end{figure*}

\subsection{Tracking} \label{sec:tracking}

\begin{algorithm}[t]
\caption{RGBD Object Tracker}
\label{alg:track}
\begin{algorithmic}[1]
\renewcommand{\algorithmicrequire}{\textbf{Input:}}
\renewcommand{\algorithmicensure}{\textbf{Output:}}
\REQUIRE RGBD video $\measurement_{t \ldots t+1}$, $\indexed{\particle}{}{}{t}$
\ENSURE $\indexed{\trajectory}{}{}{t} \from \objectIndexSpace \to \SE{3}$\\

\STATE \textsc{Project} $\from \indexed{{\particle}}{}{}{t} \mapsto \indexed{{\segmentation}}{}{}{t}$ \\
\FOR {$k = 0$ to $n$}
\STATE $\indexed{\mask}{k}{}{t+1}$ $\leftarrow$ \textsc{SiamMask}($\indexed{\mask}{k}{}{t}$, $\measurement_{t \ldots t+1}$)

\STATE Source Point Cloud $\vec{P_{src}} \leftarrow \measurement_t \odot \indexed{\mask}{k}{}{t}$
\STATE Target Point Cloud $\vec{P_{tar}} \leftarrow \measurement_{t+1} \odot \indexed{\mask}{k}{}{t+1}$

\STATE Feature displacements $\{\vec{d}_j\} \leftarrow $ \textsc{SIFT}($\vec{P_{src}}$, $\vec{P_{tar}}$)
\STATE Rigid body transform $\vec{T}$ $\leftarrow$ \textsc{JLinkage}($\{\vec{d}_j\}$)

\IF {\textsc{NumberInliers}($\{\vec{d}_j\}$, $\vec{T}$) $> \text{thres}_{SIFT}$}
\STATE $\indexed{\trajectory}{k}{}{t} \leftarrow \vec{T}$
\STATE continue
\ENDIF

\STATE $\vec{T} \leftarrow$ \textsc{ICP}($\vec{P_{src}}$, $\vec{P_{tar}}$)
\IF {\textsc{FitError}($\vec{P_{src}}$, $\vec{P_{tar}}$, $\vec{T}$) $< \text{thres}_{ICP}$}
\STATE $\indexed{\trajectory}{k}{}{t} \leftarrow \vec{T}$
\ELSE
\STATE $\indexed{\trajectory}{k}{}{t} \leftarrow \vec{I}$
\ENDIF

\ENDFOR
\RETURN $\indexed{\trajectory}{}{}{t}$
\end{algorithmic}
\end{algorithm}

Static scenes can contain ambiguities that can be challenging even for human observers, so we introduce motion into the scene to assist in differentiating object boundaries.
We use video/object tracking to compute rigid body trajectories $\trajectory \from \objectIndexSpace \to \SE{3}$ for each segment from an observation sequence $\measurement_{t \ldots t+1}$ (see Alg. \ref{alg:track}, where $\odot$ denotes element-wise multiplication).
The process consists of two steps: object mask tracking and transform estimation.

\subsubsection{Video Object Tracking}
We employ the state-of-art video object tracking algorithm SiamMask \cite{wang2019fast} to determine the correspondence of objects between the frames at $t$ and $t+1$. 
The output mask video from SiamMask will be further utilized in rigid body transformation estimation of each object. 
SiamMask requires the bounding box of a target object as input, so we first project each volumetric representation $\indexed{\particle}{}{}{t}$ onto a 2-D segmentation image by $\textsc{Project}$, then compute a bounding box for each unique label in the new segmentation image. 
For each object $\objectFull$, we apply SiamMask to the recorded video between $t$ and $t+1$ and its corresponding bounding box, generating the single-frame mask $\indexed{\mask}{k}{i}{t+1}$ of the object $\indexed{\object}{k}{i}{t+1}$.

\subsubsection{Rigid Body Transformation Estimation}
In order to estimate the motion $\trajectory$ of every object, we first compute the rigid body transformation by matching SIFT \cite{lowe2004distinctive} key points in $\indexed{\mask}{k}{i}{t}$ to those in $\indexed{\mask}{k}{i}{t+1}$. However, there are sometimes erroneously-matched features because some objects have multiple similar SIFT key points.
We thus use JLinkage \cite{toldo2008robust}, a clustering algorithm which is able to handle outliers (similar to RANSAC + Hough voting), to obtain a candidate rigid body transformation from the SIFT matches.

SIFT works well on objects with distinguishable feature points, but it works poorly on textureless objects.
As a fallback, we also compute the transform between the parts of the point cloud corresponding to $\indexed{\mask}{k}{i}{t}$ and $\indexed{\mask}{k}{i}{t+1}$ using Iterative Closest Point (ICP). We use the best fit of JLinkage or ICP to estimate $\trajectory$.
If both methods fail, a transform of identity is assumed.
This approach performs well when the tracked object is occluded by an object pushed in front of it, but struggles when the tracking failure is due to erroneous masking.

\subsection{Merging in New Information} \label{sec:fusing}

While tracking can be effective, some events cannot be represented by our transforms $\trajectory$, such as previously-unseen objects being revealed between $t$ and $t+1$, so we need another way to update our segmentation estimates to accord with new information. Related work in particle filtering has considered an analogous problem: Manifold particle filters \cite{koval2017manifold} use the measurement step to perform a projection $\projection \from \stateSpace^\comp \to \boundary \stateSpace$ carrying invalid (or highly improbable) states to a nearby valid one.
Evolutionary particle filters \cite{kwok2005evolutionary} forego the default resampling process in favor of a genetic algorithm style crossover operation to update the particle population.
We employ the first technique using \textit{free space refinement}: voxels believed to be in free space based on $\indexed{\measurement}{}{}{t}$ are cleared for all $\indexed{\predicted{\particle}}{}{i}{t}$. We also construct a quality function $\refinementQuality(\indexed{\predicted{\particle}}{}{i}{t})= {\card{\indexed{\predicted{\particle}}{}{i}{t}}_{after}}/{\card{\indexed{\predicted{\particle}}{}{i}{t}}
_{before}}$ describing the level of error corrected by refinement.

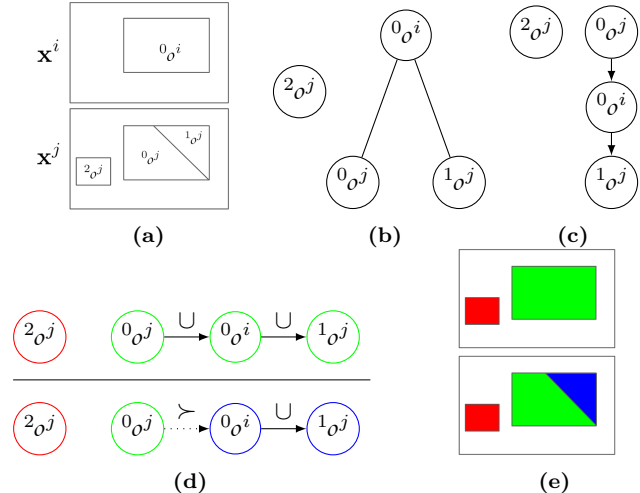
\begin{figure}[t]
    \setlength{\belowcaptionskip}{0pt}
    \centering
    \begin{subfigure}[b]{.35\linewidth}
        \includegraphics[height=2.75cm]{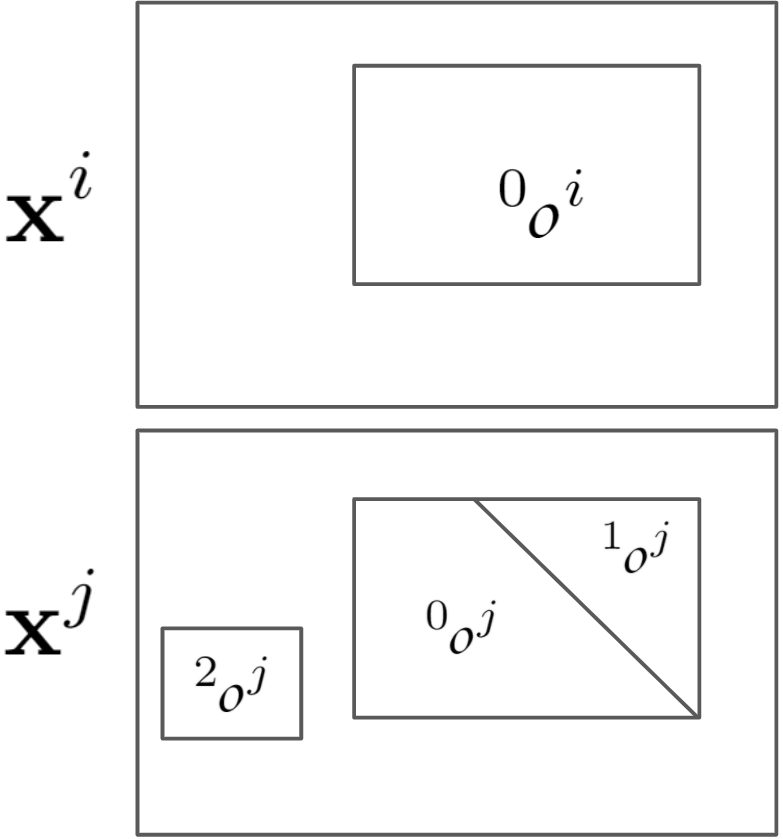}
        \caption{}
    \end{subfigure}
    \begin{subfigure}[b]{.35\linewidth}
        \begin{tikzpicture}[node distance=1cm,
                            n1/.style={circle, draw=black, inner sep=1pt, minimum size=0.1cm}]
            \node[n1] (a) {$\indexed{{\object}}{2}{j}{}$};
            \node[n1] (b) [below right of=a, yshift=-0.5cm] {$\indexed{{\object}}{0}{j}{}$};
            \node[n1] (c) [above right of=b, yshift=1.25cm] {$\indexed{{\object}}{0}{i}{}$};
            \node[n1] (d) [below right of=c, yshift=-1.25cm] {$\indexed{{\object}}{1}{j}{}$};
            \draw (b) -- (c);
            \draw (c) -- (d);
        \end{tikzpicture}
        \caption{}
    \end{subfigure}
    \begin{subfigure}[b]{.2\linewidth}
        \begin{tikzpicture}[->,>=latex,node distance=1cm,
                            n1/.style={circle, draw=black, inner sep=1pt, minimum size=0.1cm}]
            \node[n1] (a) {$\indexed{{\object}}{2}{j}{}$};
            \node[n1] (b) [right of=a] {$\indexed{{\object}}{0}{j}{}$};
            \node[n1] (c) [below of=b] {$\indexed{{\object}}{0}{i}{}$};
            \node[n1] (d) [below of=c] {$\indexed{{\object}}{1}{j}{}$};
            \draw (b) -- (c);
            \draw (c) -- (d);
        \end{tikzpicture}
        \caption{}
    \end{subfigure}

    \begin{subfigure}[b]{.55\linewidth}
        \begin{tikzpicture}[->,>=latex,node distance=1.3cm,
                            n1/.style={circle, draw=red, inner sep=1pt, minimum size=0.1cm},
                            n2/.style={circle, draw=green, inner sep=1pt, minimum size=0.1cm},
                            n3/.style={circle, draw=blue, inner sep=1pt, minimum size=0.1cm}]
            \node[n1] (a) {$\indexed{{\object}}{2}{j}{}$};
            \node[n2] (b) [right of=a] {$\indexed{{\object}}{0}{j}{}$};
            \node[n2] (c) [right of=b] {$\indexed{{\object}}{0}{i}{}$};
            \node[n2] (d) [right of=c] {$\indexed{{\object}}{1}{j}{}$};
            \draw (b) -- node[midway, above] {$\union$} (c);
            \draw (c) -- node[midway, above] {$\union$} (d);
        \end{tikzpicture}

        \par\noindent\rule[2mm]{\textwidth}{0.3pt}

        \begin{tikzpicture}[->,>=latex,node distance=1.3cm,
                            n1/.style={circle, draw=red, inner sep=1pt, minimum size=0.1cm},
                            n2/.style={circle, draw=green, inner sep=1pt, minimum size=0.1cm},
                            n3/.style={circle, draw=blue, inner sep=1pt, minimum size=0.1cm}]
            \node[n1] (a) {$\indexed{{\object}}{2}{j}{}$};
            \node[n2] (b) [right of=a] {$\indexed{{\object}}{0}{j}{}$};
            \node[n3] (c) [right of=b] {$\indexed{{\object}}{0}{i}{}$};
            \node[n3] (d) [right of=c] {$\indexed{{\object}}{1}{j}{}$};
            \draw[dotted] (b) -- node[midway, above] {$\succ$} (c);
            \draw (c) -- node[midway, above] {$\union$} (d);
        \end{tikzpicture}
        \caption{}
    \end{subfigure}
    \hfill
    \begin{subfigure}[b]{.3\linewidth}
        \includegraphics[height=2.75cm]{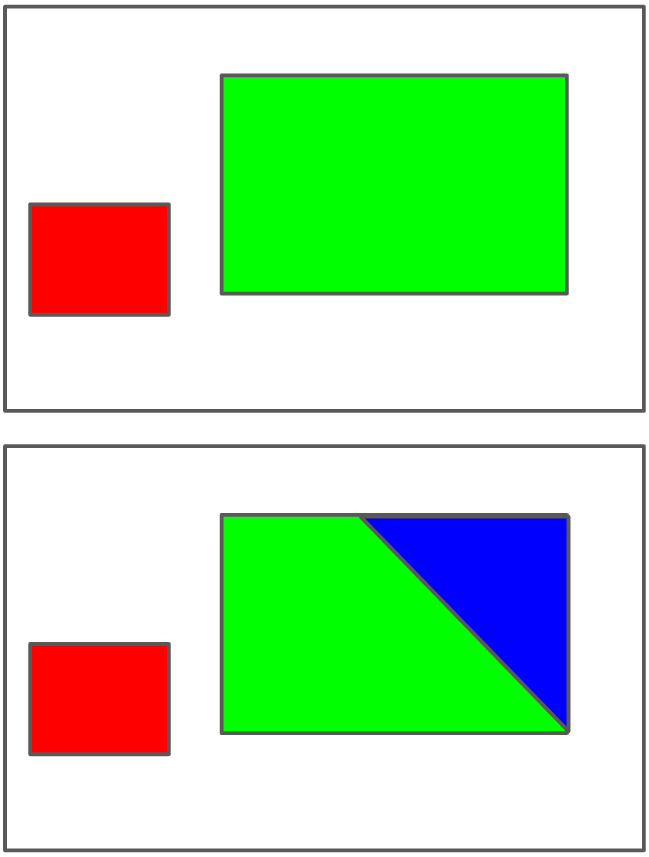}
        \caption{}
    \end{subfigure}
    \setlength{\belowcaptionskip}{-10pt}
    
\caption{An illustration of the merging process in 2D. (a) Two segmentations to be merged. (b) The conflict graph. (c) A randomly sampled topological order, in which parent nodes will overwrite child nodes. (d) Two different merge samples. Operator $A \succ B$ here means that segments $A$ and $B$ will not be merged, and labels from $A$ will have the higher precedence according to the ordering. Nodes of the same color have been merged ($\union$). (e) The two resulting segmentations.}
\label{fig:merge}
\end{figure}

\begin{figure*}[t]
    \centering
    \begin{subfigure}[b]{.19\linewidth}
        \includegraphics[width=\linewidth]{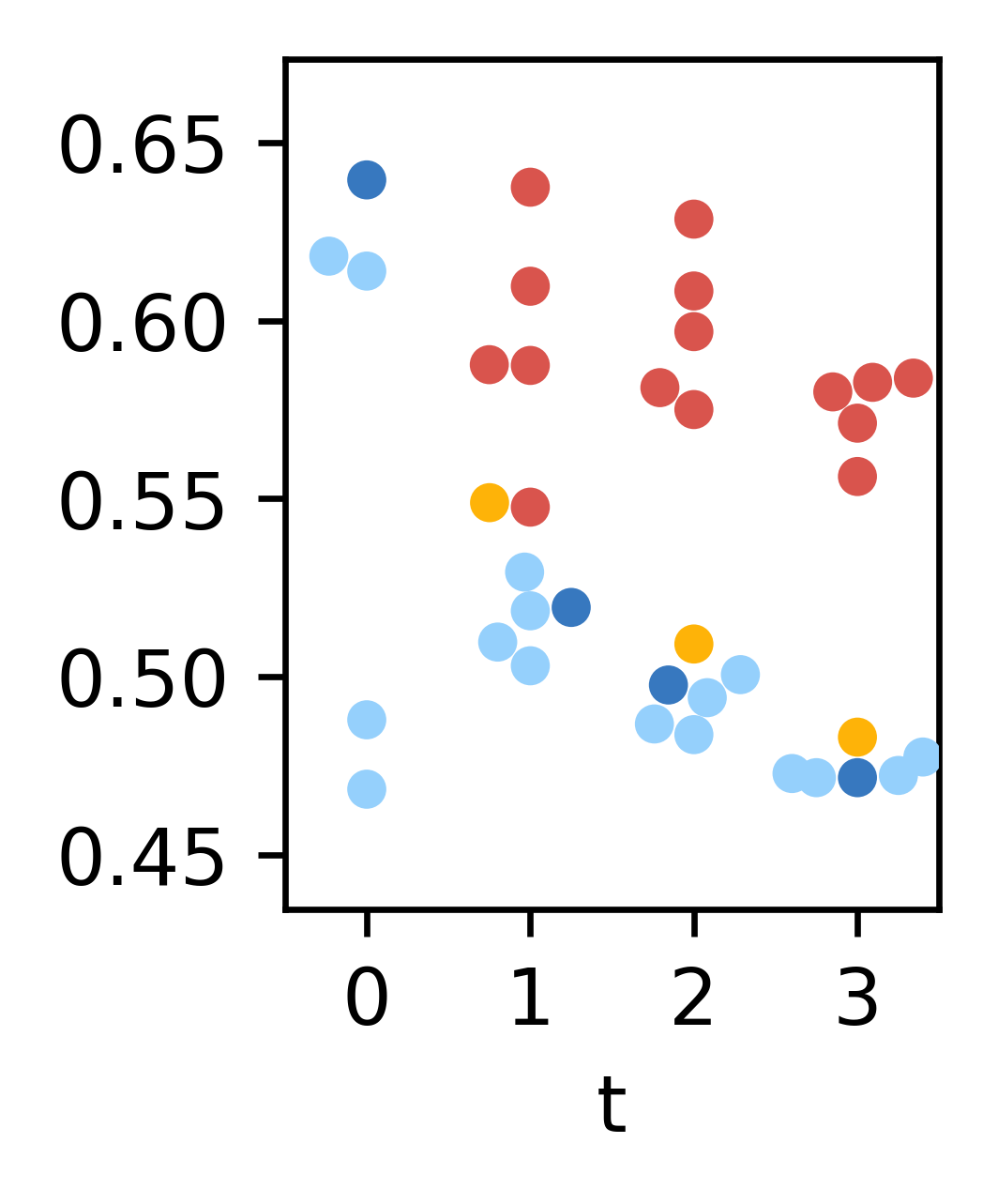}
    \end{subfigure}
    \begin{subfigure}[b]{.19\linewidth}
        \includegraphics[width=\linewidth]{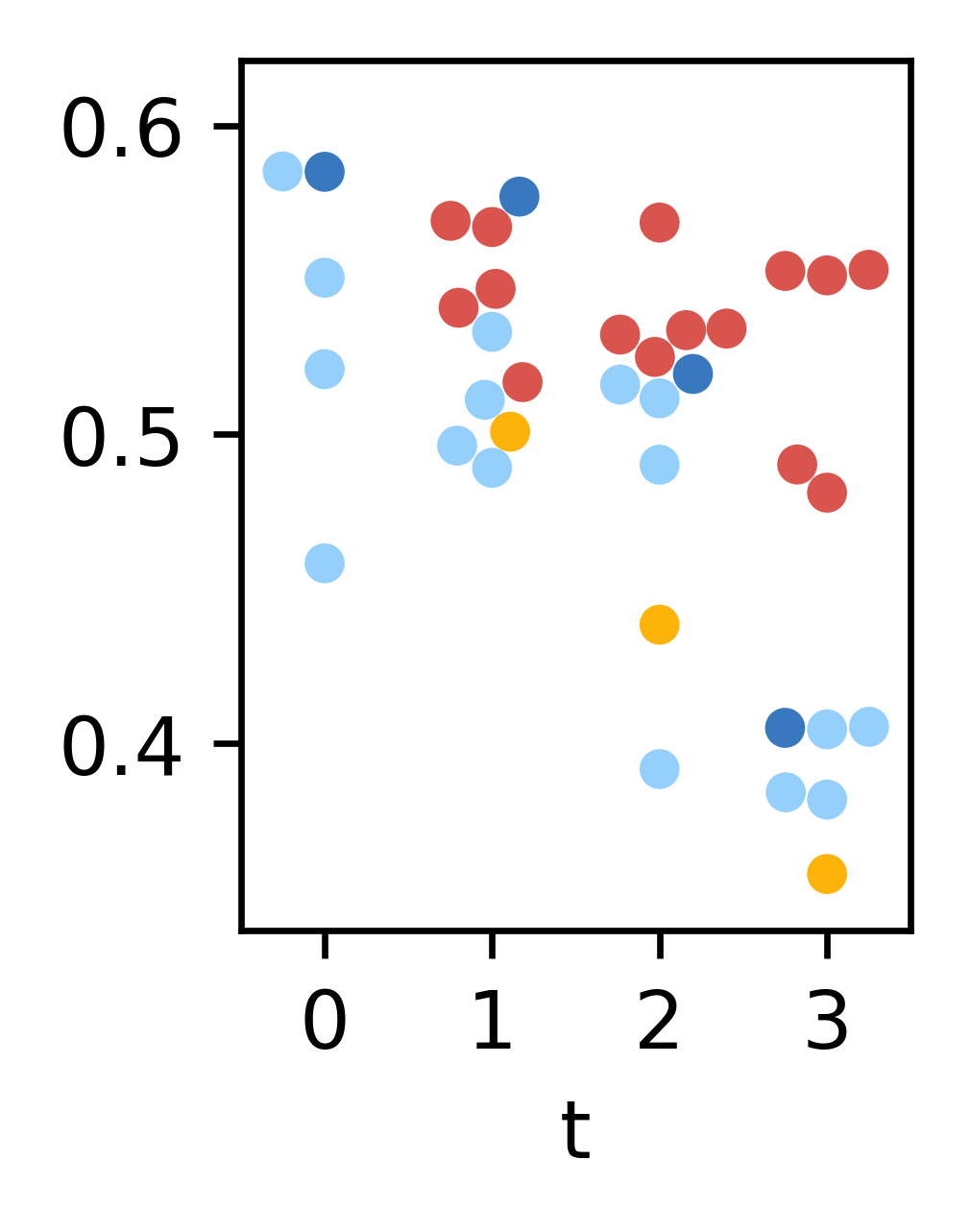}
    \end{subfigure}
    \begin{subfigure}[b]{.19\linewidth}
        \includegraphics[width=\linewidth]{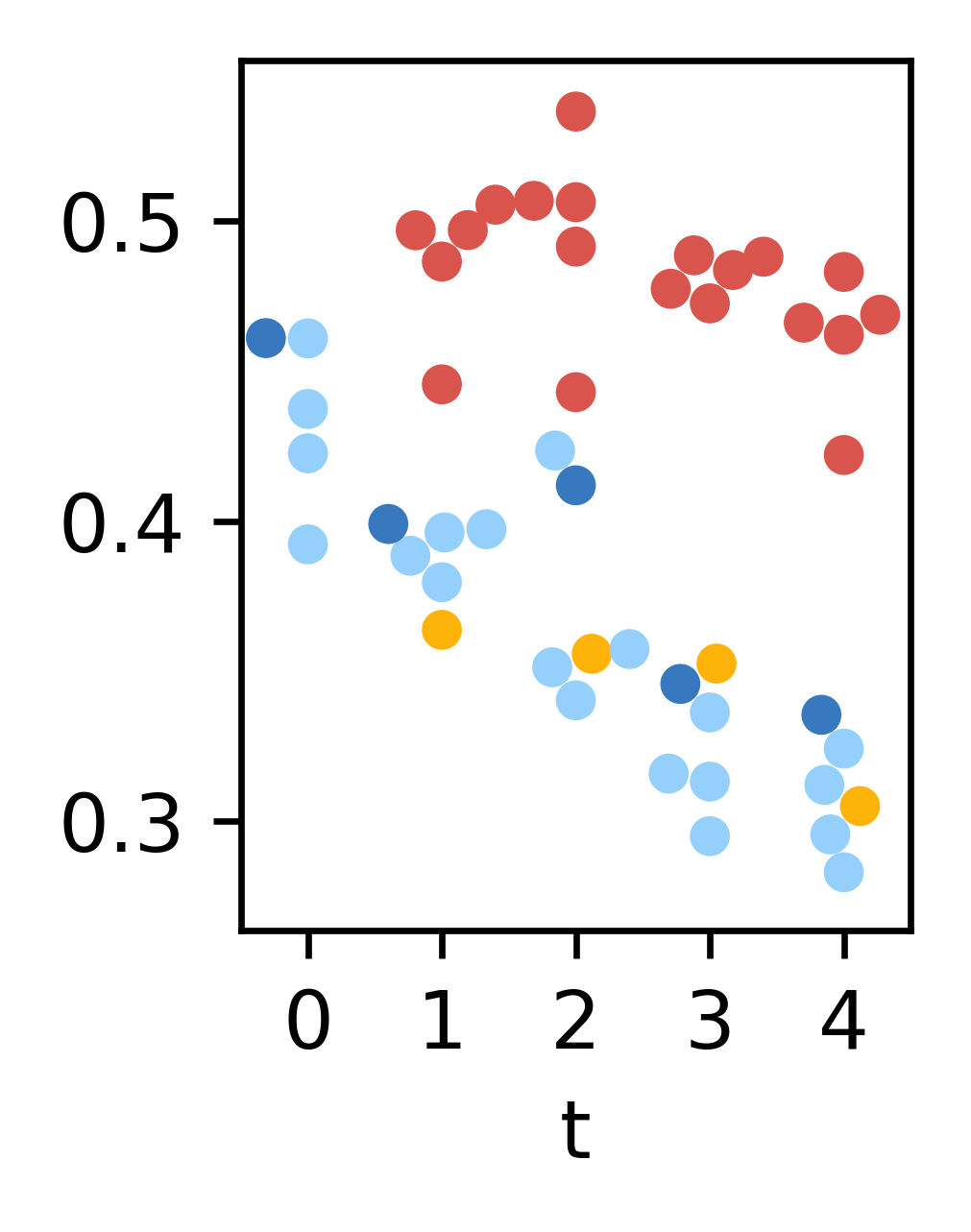}
    \end{subfigure}
    \begin{subfigure}[b]{.19\linewidth}
        \includegraphics[width=\linewidth]{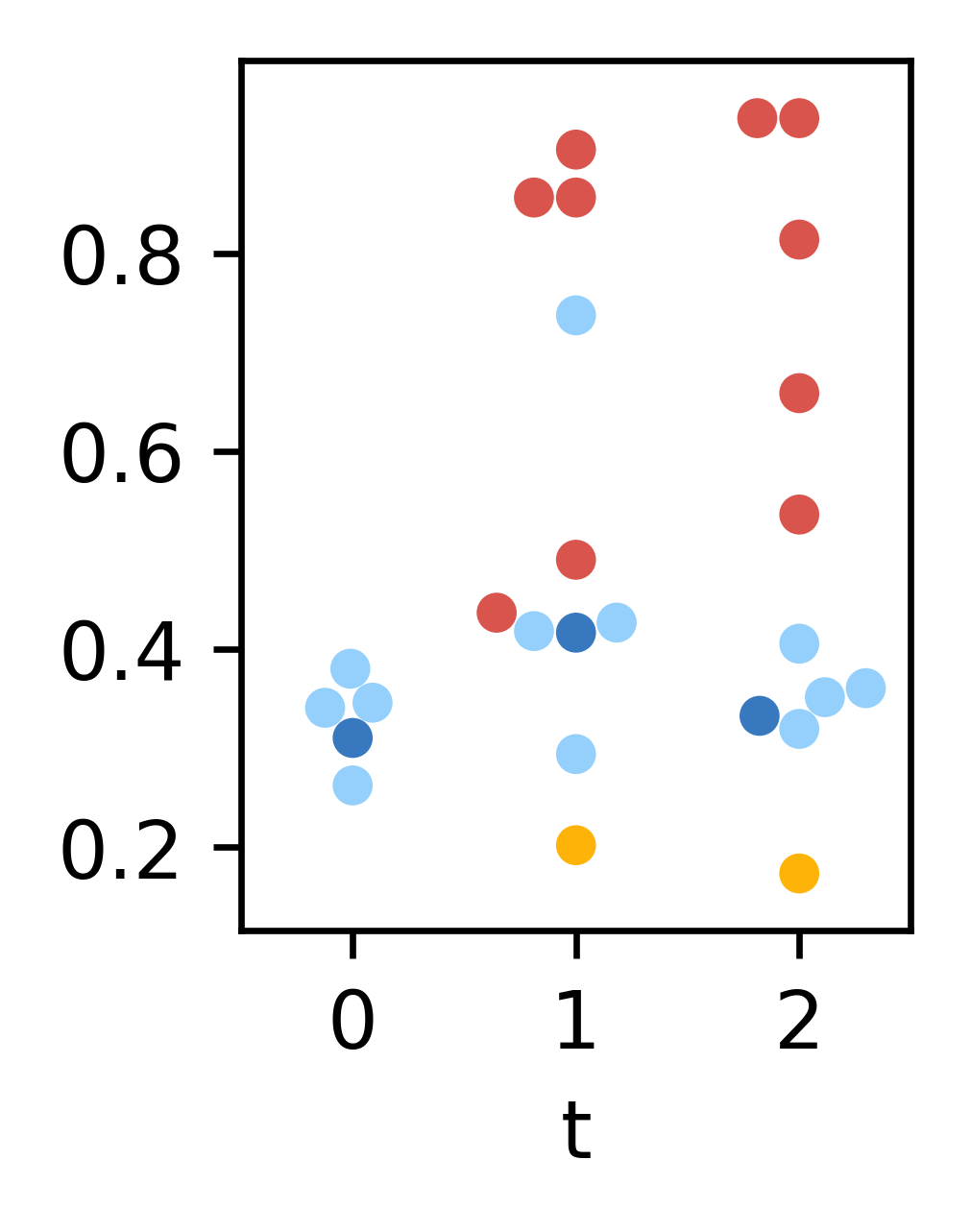}
    \end{subfigure}
    \begin{subfigure}[b]{.19\linewidth}
        \includegraphics[width=\linewidth]{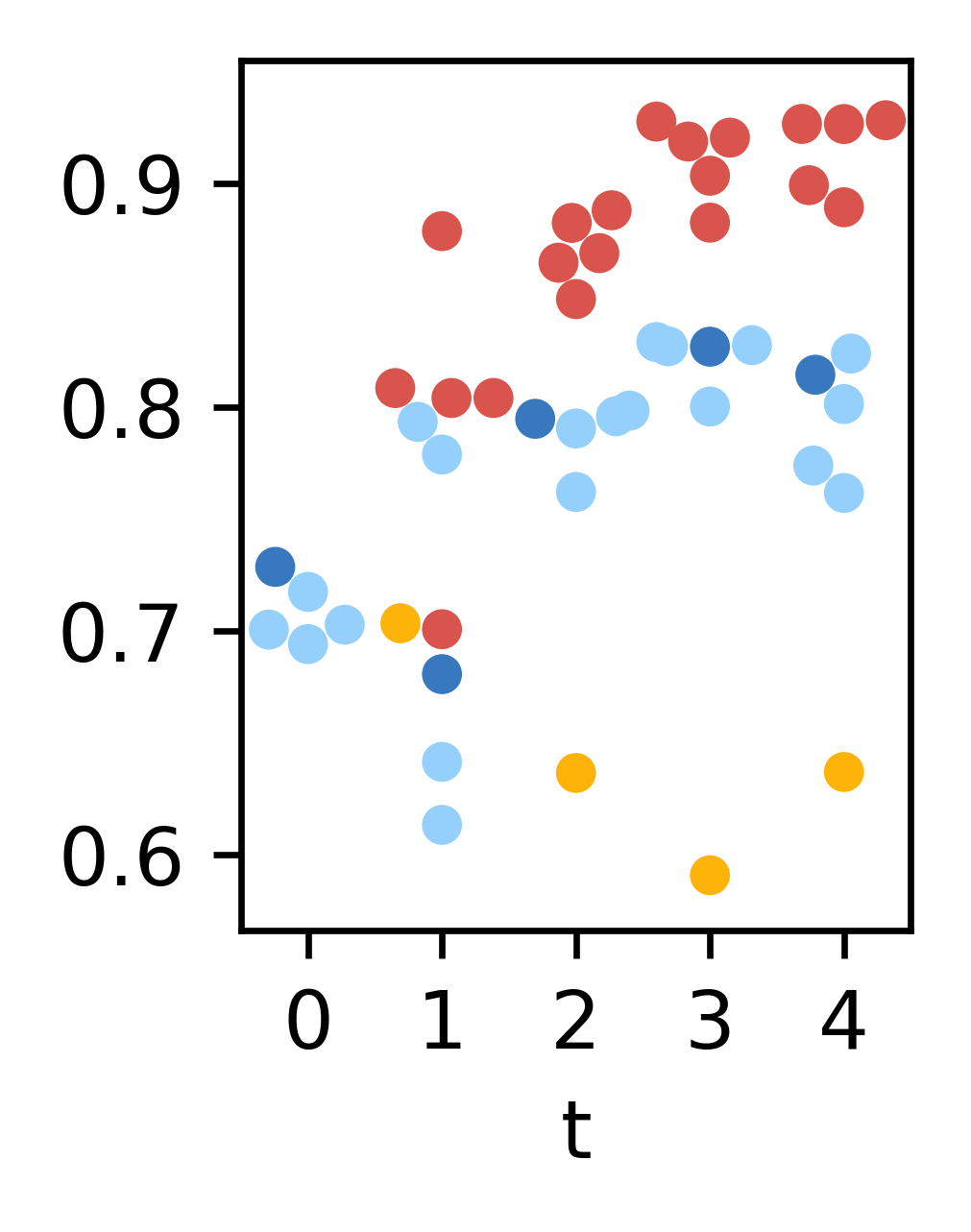}
    \end{subfigure}
\caption{Left to right: coverage quality $\quality$ of hypotheses from various algorithms vs. ground truth for experiments S1, S2, S3, R1, and R2. Dark blue: single-frame segmentation \cite{price2019inferring} baseline; Light blue: segmentation samples $\sampled{\particle}$; Orange: tracking-only baseline \cite{wang2019fast}; Red: set of hypotheses $\particle$ produced by MST. Higher is better. (Note the recovery from the sampler's oversegmentation bias in R1-2.)}
\label{fig:results}
\end{figure*}

We propose a novel approach to accommodate situations where objects have appeared in the scene, inspired by \cite{kwok2005evolutionary}. We first generate a new set of samples $\indexed{\sampled{\particle}}{}{i}{t}$ which we then combine with the $\indexed{\predicted{\particle}}{}{j}{t}$ generated by tracking to form the final estimate $\indexed{\particle}{}{}{t}$. 
The procedure is as follows:
For a pair $(i, j)$ of segmentations between the predicted and newly sampled populations, we generate a conflict graph $G(n, e)$ where nodes $n$ represent objects in $\indexed{\sampled{\particle}}{k}{i}{t}$ or $\indexed{\predicted{\particle}}{\ell}{j}{t}$, and edges $e$ with weights $\indexed{w}{}{}{e}$ representing the voxel IOU between the objects.
Thus, objects that are newly occluded or disoccluded will have no conflicts, and will be inserted directly into the resulting state.

For each connected component in $G$, we need to decide whether to merge the competing objects, or whether to separate them with only one object ``winning'' for a given voxel. Since it is unclear which hypothesis to favor when merging/separating, we sample over the possibilities to preserve diversity in our set of hypotheses: We induce a precedence order by applying a random edge orientation and topological ordering to the component, then sample merges from node $n_1$ to its parent node $n_2$ with probability $\indexed{w}{}{}{e}{\card{n_2}}/{\card{n_1}^2}$, so that we can get similar probability for merging and separating.
A consensus-thresholded mode filter, which replaces voxels with the most frequently occurring voxel value selected from a certain window size, is applied to postprocess the state, cleaning up cases where only the periphery of an object survived the merger (usually due to a small misalignment between the prediction and new sample).
Applying $\eta$ samples of this procedure to all pairs $(i, j)$ produces a population much larger than the desired number of hypotheses, so a weight 
$\indexed{w}{}{m}{t+1} = 
\indexed{w}{}{j}{t} 
\indexed{\scenecutWeight}{}{i}{t+1}
\refinementQuality(\indexed{\predicted{\particle}}{}{j}{t+1})^\lambda
$
is used to downsample the population back down to the desired size, with selection probability for object $m$ $\propto \indexed{w}{}{m}{t+1}$. The hyper-parameter $\lambda$ represents the relative trust in tracking results.

\section{Experiments and Results} \label{sec:experiments}
To evaluate the performance of MST it would be ideal to compare to methods that perform volumetric segmentation in changing scenes.
The closest approach we are aware of is MID-Fusion \cite{xu2019midfusion}, a volumetric dynamic SLAM algorithm, but we are unaware of an open-source implementation.
We test in three simulation and two real-world tabletop scenes. In simulation all approaches are evaluated by comparing to the volumetric ground truth according to the $\quality$ function above (\cref{eq:quality_fn}), while the real-world results are evaluated with respect to a hand-labeled 2D image segmentation ground truth.
\Cref{fig:experiments} shows the initial conditions of all experiments.
We use voxels of size 1cm, and a workspace of ~1m x 1m x 0.5m containing 2-20 objects.
For algorithm parameters, we set sampling $\sigma=0.25v(c^*)^2$, $\lambda = 3$, $\hypothesisCount =5$, $\text{thres}_{SIFT} = 5$ and $\text{thres}_{ICP} = 1$mm in all experiments.
See the accompanying video for visualizations of results.

\subsection{Simulation Experiments}
To evaluate the system with known 3D ground truth segmentation results, we created simulated scenes in Gazebo, then used a simulated robot to push and pull objects.
The robot performed 3-4 manipulation actions, generated via the technique described in \cite{price2019inferring}, pausing to observe the scene between each.
Experiments S1, S2, and S3 demonstrate increasing degrees of clutter, with duplicated objects contacting and occluding one another.

\Cref{fig:results} shows the quality of segmentations over time, with the results of our method clearly outperforming baselines.
Of particular note is the generally decreasing quality trends in S1 and S2.
Due to the difficulty of simulating grasping in Gazebo and the robot's limited dextrous workspace, many of the sampled robot actions push the objects into tighter configurations with greater occlusion and increased segmentation difficulty.
By retaining some of the information from earlier, less ambiguous scenes, our approach is able to maintain higher quality over time.

\subsection{Real-world Experiments}
There are significant sensing differences between simulation and physical environments, particularly involving sensor noise, lighting, and object texture, so we have also evaluated our system on real tabletop environments utilizing objects from the YCB dataset \cite{calli2015benchmarking}.
Experiment R1, detailed in \cref{fig:realworld}, is an illustrative example showing how performing the estimation in 3D can help recover from poor initial segmentations.
In particular, R1 shows that the un-tuned segmentation is biased towards oversegmentation of the initial scene, but MST is able to recover from this by incorporating the object motion.
Experiment R2 shows a more challenging and realistic scene in which objects are pushed from one cluster to another.
Performance results are shown in \cref{fig:results}: MST strongly outperforms the baselines because of its ability to fuse hypotheses in 3D and retain them in the presence of occlusion.

Currently, the pipeline implementation has significant runtime, requiring $\sim$1 hour to process a sequence of images from one of the experiments described above.
The primary bottleneck is currently serialized calls to the dense video tracking, which must be done for every segment in every hypothesis at every time step.
The redundant calls can be reduced and they could be parallelized with additional engineering effort, and there are many opportunities for GPU implementations of the occupancy operations.

\section{Conclusions} \label{sec:conclusions}
In this work, we have shown how our method for multi-hypothesis volumetric estimation can outperform single-frame scene segmentation and tracking-propagated segmentation, particularly in cluttered manipulation scenarios.
To do so, we have introduced novel techniques for sampling different segmentations and for combining them after perceived motions.
Experiments in simulated and real environments show that this technique is promising for tabletop manipulation in challenging scenes.
However, there remain significant opportunities to improve the system.
As with any serialized data pipeline, individual improvements to the accuracy of any one stage will improve the overall performance, just as egregious errors will derail an otherwise reasonable segmentation hypothesis. Thus advances in image segmentation, shape completion, and 3D registration will boost the accuracy of the system overall.
Second, there are significant engineering gains to be made by improving the parallelism at all levels of the system, from voxels to hypotheses.
Third, improved use of physical understanding of the scene could allow for more accurate transition predictions, enhancing our ability to resolve voxel conflicts.
Despite these avenues for future improvement, we feel these results provide a case in favor of explicitly modeling and propagating uncertainty in sequential instance segmentation.

{
\bibliographystyle{unsrt}
\bibliography{egbib}
}

\end{document}